\documentclass[runningheads]{llncs}

\usepackage[year=2026]{eccv}

\usepackage{eccvabbrv}

\usepackage{amsmath,amssymb}
\usepackage{graphicx}
\usepackage{booktabs}
\usepackage{tikz}
\usepackage{multirow}

\usepackage[accsupp]{axessibility}

\usepackage{hyperref}
\usepackage{wrapfig}

\newcommand{\mat}[1]{\mathbf{#1}}
\renewcommand{\vec}[1]{\mathbf{#1}}
\renewcommand{\skew}[1]{[#1]_{\times}}

\begin{document}

\title{Rolling Shutter Relative Pose\\Estimation Made Practical}

\titlerunning{RS Relative Pose Made Practical}

\author{Daniel Barath\inst{1,2,3}}
\authorrunning{D. Barath}
\institute{ETH Z\"urich, Computer Vision and Geometry Group,
  \email{dbarath@ethz.ch}
  \and Google Zurich \and  HUN-REN SZTAKI, Budapest, Hungary}

\maketitle

\begin{abstract}
Rolling shutter (RS) cameras equip virtually all consumer devices,
yet RS-aware relative pose estimation has remained impractical:
the state-of-the-art solver requires a minimum of 20 point
correspondences, making RANSAC-based robust estimation
prohibitively expensive due to the exponential dependence
of the iteration count on the sample size.
We make RS relative pose estimation practical by introducing
affine correspondences (ACs) into the RS two-view geometry.
We derive novel \emph{RS-corrected affine constraints}
that account for the coupling between point perturbations
and the row-dependent essential matrix,
providing two equations per correspondence
beyond the standard epipolar constraint.
Building on these constraints, we develop a linearized algebraic
solver that estimates pose and RS motion from only
7 ACs.
The solver exploits the physical smallness of RS parameters
to linearize the constraints, eliminates the 12 RS unknowns
via null-space projection, and solves the remaining degree-20
system via action matrices in 1.2\,ms.
On the TUM RS benchmark, our method achieves
the best pose and RS parameter accuracy among
all tested methods and, uniquely among RS solvers,
provides accurate translational velocity estimates --
which are poorly conditioned from point correspondences
alone due to a $\vec{v}$-$\vec{t}$ coupling.
On the global-shutter EuRoC MAV dataset, the solver achieves comparable accuracy to the standard 5-point algorithm,
demonstrating that it generalizes well to the GS setting.
Code is at \url{https://github.com/danini/rolling_shutter_made_practical}.
\end{abstract}

\begin{figure}[t]
\centering
\includegraphics[width=0.99\linewidth]{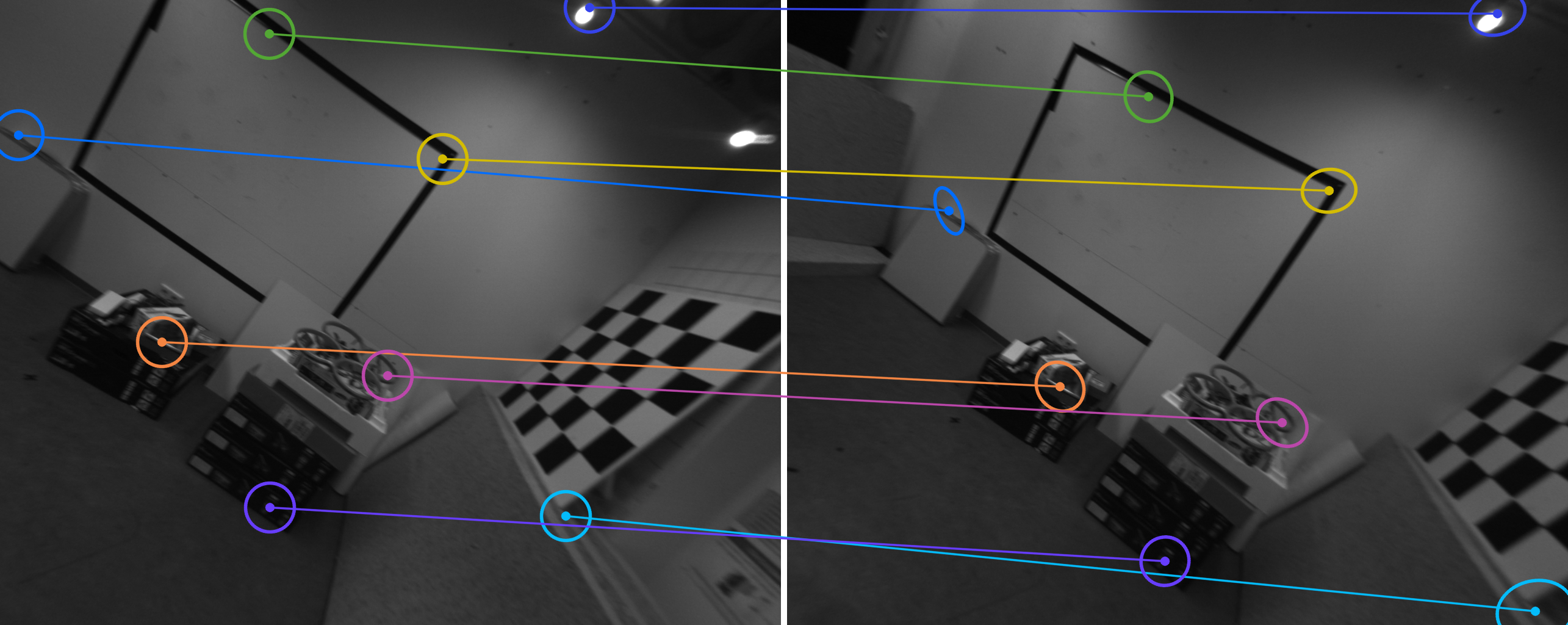}

\caption{\textbf{Affine correspondences on a rolling shutter image pair}
from the TUM-RS dataset~\cite{Schubert2018}.
Circles in the left image and ellipses in the right image represent the
local affine frames of 8 matched features.
The camera rotates ${\sim}8.5$° during each frame's readout,
producing visible RS distortion.
On this pair, a standard GS 5-point solver~\cite{Nister2004} yields
rotation/translation errors of $0.64$°/$4.33$°;
our RS-aware 7-AC solver reduces these to $0.20$°/$2.54$°
while additionally recovering the RS motion parameters.}
\label{fig:teaser}

\end{figure}

\section{Introduction}\label{sec:intro}

Rolling shutter (RS) cameras equip virtually all consumer
devices -- smartphones, tablets, action cameras, and drones.
Unlike global shutter (GS) sensors that expose all pixels
simultaneously, RS sensors read out the image row by row.
When the camera moves during the readout interval -- typically
15--30\,ms for consumer CMOS sensors -- each row captures the
scene from a slightly different viewpoint.
This sequential exposure invalidates the standard pinhole
model that underpins most multi-view geometry
methods~\cite{Meingast2005}: the epipolar constraint,
which normally defines a constant essential matrix across
the image, becomes row-dependent.

Relative pose estimation is fundamental to structure from
motion (SfM)~\cite{Hedborg2012}, visual odometry, SLAM,
and augmented reality.
These systems typically treat the essential matrix $\mat{E}$
as constant across the image.
On RS images, this introduces a systematic bias that grows
with camera velocity and readout time, and in severe cases
causes bundle adjustment to diverge or converge to incorrect
reconstructions.
As RS cameras continue to dominate consumer and industrial
markets, developing RS-aware geometric methods is essential.

Under the first-order RS model of Meingast et al.~\cite{Meingast2005},
the RS relative pose problem has 17 degrees of freedom:
5 for the relative pose $(\mat{R}, \vec{t})$ plus 12 for the
per-camera angular and translational velocities
$(\vec{\omega}_k, \vec{v}_k)$, $k=1,2$.
Dai et al.~\cite{Dai2016} showed that a minimum of
\textbf{20 PCs} is required -- far more than the 5 PCs
of GS estimation~\cite{Nister2004}.
This gap has practical consequences.
The expected RANSAC~\cite{Fischler1981} iteration count scales as
$\eta^{-k}$ (inlier ratio $\eta$, sample size $k$): at $50\%$ inliers,
$k\!=\!5$ needs $32$ iterations but $k\!=\!20$ over a million -- making
the 20-PC solver~\cite{Dai2016} prohibitive and motivating smaller samples.

\emph{Affine correspondences} (ACs) provide a natural path
toward this goal.
An AC augments a point match with the local $2 \times 2$ affine
transformation between image
patches~\cite{Bentolila2014,Barath2018},
yielding 3 equations per correspondence
(1 epipolar + 2 affine).
In the GS setting, this tripling of constraints enables
essential matrix recovery from just
2 ACs~\cite{Barath2018} instead of 5 PCs,
and homographies from 2 ACs~\cite{BarathHomography2017}
instead of 4 PCs.
ACs are routinely available from affine-covariant
detectors~\cite{Mikolajczyk2004} and learned
methods~\cite{Mishkin2018,sun2025learning} at negligible additional cost.

We make RS relative pose estimation practical by bringing ACs
to the problem for the first time.
We derive \emph{RS-corrected affine constraints} that account
for the coupling between point perturbations and the
row-dependent essential matrix -- a coupling absent in the
GS formulation -- and build a linearized algebraic solver
that estimates pose and RS parameters from
7 ACs, reducing RANSAC samples from 20 to 7
and iterations from millions to hundreds.
On real-world data, we achieve the best pose and RS parameter
estimation accuracy among all tested methods,
demonstrating that RS-aware relative pose estimation is now
practical with affine correspondences.
Our key contributions are:
\begin{itemize}
\item We derive \emph{RS-corrected affine correspondence
  constraints} that account for the row-dependent essential
  matrix -- a coupling absent in the GS formulation.
  Each AC yields 3 equations (1 epipolar + 2 affine),
  so 7 ACs provide 21 equations for the 17 unknowns
  of the joint pose + RS problem.
\item We develop a 7-AC algebraic solver with
  algebraic degree 20, based on action matrix
  methods, that makes RS-aware RANSAC practical:
  the sample size reduction from 20 to 7
  decreases the expected RANSAC iterations by orders
  of magnitude
  (e.g., from $10^6$ to $\sim$600 at 50\% inliers).
\item We show that the proposed method
  achieves the best pose and RS estimation accuracy on the TUM-RS
  benchmark among RS and GS baselines, and
  substantially improves translational velocity estimation --
  which is poorly conditioned from PCs alone
  due to a $\vec{v}$-$\vec{t}$ coupling~\cite{Dai2016}.
  On the global-shutter EuRoC MAV dataset~\cite{Burri2016},
  the solver generalizes without degradation.
\end{itemize}
%

\section{Related Work}\label{sec:related}

\vspace{1mm}\noindent\textit{RS camera modeling and pose estimation.}
Meingast et al.~\cite{Meingast2005} introduced the first-order RS model
in which each scanline's pose is linearly interpolated between
a reference rotation and translation, adding six motion parameters
(angular and translational velocity) per frame.
This model has been validated across consumer
cameras~\cite{Albl2015,Dai2016,Albl2020} and remains the standard
choice when the readout-time rotation is moderate
($\lesssim 10^\circ$).
For absolute pose, Albl et al.~\cite{Albl2015,Albl2016}
developed Gr\"obner-basis minimal solvers under the linearized
RS model (6-point, or 5-point with known gravity);
Saurer et al.~\cite{Saurer2015} proposed an alternative formulation;
and Albl et al.~\cite{Albl2020} later unified these results.
For relative pose, Dai et al.~\cite{Dai2016} derived the RS
epipolar geometry and showed that the linearized two-view problem
requires 20 PCs (44 for the quadratic model).
The $\vec{v}$-$\vec{t}$ coupling between the translational velocity
and the baseline direction had been identified earlier by
Ait-Aider and Berry~\cite{AitAider2009}.
Zhuang et al.~\cite{Zhuang2017} used differential homographies for
RS-aware SfM and rectification, Lee et al.~\cite{Lee2019} a gyroscope
to reduce the unknowns, and Hedborg et al.~\cite{Hedborg2012} extended
bundle adjustment to RS.
Hahn et al.~\cite{Hahn2025} recently gave an algebraic
analysis of \emph{order-one} RS cameras, under a model in which
each world point is imaged at most once -- distinct from the
first-order model used here, where a point may be observed
across several rows.
Concurrently, Hruby and Pollefeys~\cite{Hruby2025} reduce the
correspondence count via a single-scanline formulation using line
correspondences -- a different input modality.
A common theme is that RS solvers need far more correspondences than
their GS counterparts, making RANSAC expensive; our solver reduces the
RS sample to 7 ACs via affine-correspondence constraints.

\vspace{1mm}\noindent\textit{Affine correspondences in geometry estimation.}
An AC augments a point match with the local $2\times2$ affine
transformation between patches, giving richer per-correspondence
constraints.
Differentiating the epipolar constraint yields two extra equations
per AC~\cite{Bentolila2014}, which Perdoch et al.~\cite{Perdoch2009}
and Barath and Hajder~\cite{Barath2018} used to recover the essential
matrix from just 2 ACs; homographies follow from 2
ACs~\cite{BarathHomography2017}, and later work consolidated the
theory across model types~\cite{Raposo2016,BarathTPAMI2023}.
ACs are widely available from affine-covariant
detectors~\cite{Mikolajczyk2004} and learned
methods~\cite{Mishkin2018,sun2025learning} at negligible cost,
but have not been applied to rolling-shutter geometry -- the gap we
bridge.

\vspace{1mm}\noindent\textit{Minimal solvers and RANSAC.}
Nist\'er's 5-point algorithm~\cite{Nister2004} is the standard GS
essential-matrix solver, and automatic Gr\"obner-basis
generators~\cite{Kukelova2008} extended such techniques to RS
settings~\cite{Albl2015,Albl2016}.
A problem's algebraic complexity (its number of solutions) governs
practicality -- our joint 7-AC solver has only 20 -- and the
$\eta^{-k}$ cost of RANSAC~\cite{Fischler1981} and its
variants~\cite{Chum2003,BarathGCRANSAC2018,Barath2020MAGSACpp}
in sample size $k$ makes reducing $k$ (here from 20 to 7) decisive.

\begin{figure}[t]
    \centering
    \includegraphics[width=0.99\linewidth]{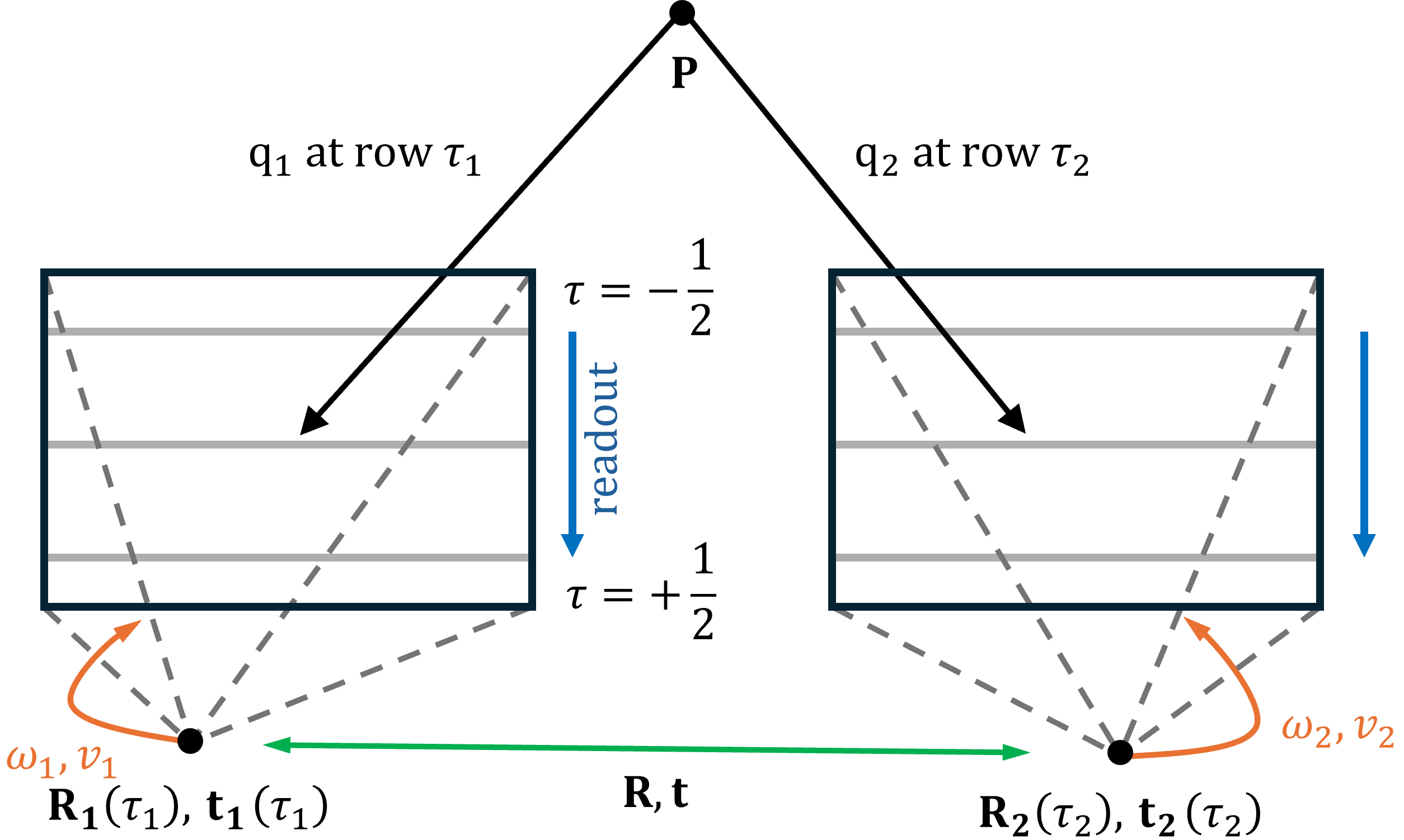}
    
    \caption{\textbf{Rolling shutter two-view geometry.}
    A 3D point $\mathbf{P}$ is observed at rows $\tau_1$ and $\tau_2$
    in two cameras whose poses vary during readout.
    Each camera is parameterized by a reference pose
    ($\mat{R}$, $\vec{t}$) and per-frame angular and
    translational velocities ($\vec{\omega}_k$, $\vec{v}_k$),
    yielding 17 degrees of freedom in total.}
    
    \label{fig:visual}
\end{figure}

\section{Geometric Background}\label{sec:background}

\noindent\textbf{Rolling Shutter Camera Model.}
A rolling shutter camera reads out the sensor row by row;
if the camera moves during the 15--30\,ms readout,
each row is captured from a different viewpoint (\cref{fig:visual}).
We adopt the standard first-order RS
model~\cite{Meingast2005,Dai2016,Albl2020}.
Let $\mat{R} \in \mathrm{SO}(3)$ and $\vec{t} \in \mathbb{R}^3$,
$\|\vec{t}\| = 1$, be the relative rotation and translation
at the reference rows, and let
$\vec{\omega}_k, \vec{v}_k \in \mathbb{R}^3$ ($k=1,2$)
denote the angular and translational velocities during readout
(absorbing the readout time $t_\mathrm{ro}$).
For a point whose pixel row is $y$, let
$\tau = (y - h/2)/h \in [-\tfrac{1}{2}, \tfrac{1}{2}]$
be its scalar normalized row coordinate;
the per-scanline pose is
\begin{align}
  \tilde{\mat{R}}(\tau_1, \tau_2)
    &= \bigl(\mat{I} + \tau_2 \skew{\vec{\omega}_2}\bigr)\,
       \mat{R}\,
       \bigl(\mat{I} - \tau_1 \skew{\vec{\omega}_1}\bigr),
       \label{eq:Rtilde} \\
  \tilde{\vec{t}}(\tau_1, \tau_2)
    &= \bigl(\vec{t} + \tau_2\,\vec{v}_2\bigr)
       - \tau_1\,\tilde{\mat{R}}\,\vec{v}_1,
       \label{eq:ttilde}
\end{align}
where $\skew{\cdot}$ denotes the skew-symmetric matrix,
$\tau_1, \tau_2 \in \mathbb{R}$ are the scalar normalized row
coordinates of the point in images 1 and 2, respectively,
and $\mat{I}$ is the identity.
The row-dependent essential matrix is
\begin{equation}\label{eq:Etilde}
  \tilde{\mat{E}}(\tau_1, \tau_2) = \skew{\tilde{\vec{t}}(\tau_1, \tau_2)}\;\tilde{\mat{R}}(\tau_1, \tau_2).
\end{equation}

The unknowns of the RS two-view problem decompose into
the \emph{pose parameters} $(\mat{R}, \vec{t})$ (5 degrees of freedom,
as translation is known only up to scale)
and the \emph{RS motion parameters}
$\vec{\theta} = (\vec{\omega}_1, \vec{v}_1, \vec{\omega}_2, \vec{v}_2) \in \mathbb{R}^{12}$,
giving \textbf{17 degrees of freedom} in total for the entire
system~\cite{Dai2016}.

\vspace{1mm}\noindent\textbf{Point Correspondences in GS and RS Cameras.}
In the global shutter (GS) case ($\vec{\theta} = \vec{0}$),
$\mat{E} = \skew{\vec{t}}\,\mat{R}$ is constant and a point pair
$(\vec{q}_1, \vec{q}_2)$ satisfies
\begin{equation}\label{eq:epipolar_gs}
  \vec{q}_2^\top \mat{E}\, \vec{q}_1 = 0,
\end{equation}
recoverable from 5 PCs~\cite{Nister2004}.
In the RS case, each pair has its own
$\tilde{\mat{E}}(\tau_1, \tau_2)$ and the constraint becomes
\begin{equation}\label{eq:epipolar}
  \vec{q}_2^\top\, \tilde{\mat{E}}(\tau_1, \tau_2)\, \vec{q}_1 = 0.
\end{equation}
Each PC still yields one equation, but the unknowns now include
the 12 RS parameters $\vec{\theta}$. The known-pose subproblem
requires 12~PCs; the joint problem requires
\textbf{20~PCs}~\cite{Dai2016} (44 for the quadratic RS model).

\vspace{1mm}\noindent\textbf{Affine Correspondences in GS Cameras.}
An affine correspondence (AC) augments a point match $(\vec{q}_1, \vec{q}_2)$
with the local affine map
$\mat{A}_c = \partial \vec{q}_2 / \partial(q_{1,x}, q_{1,y}) \in \mathbb{R}^{3 \times 2}$,
available from affine-covariant
detectors~\cite{Mikolajczyk2004,Mishkin2018}.
Let $\vec{a}_u, \vec{a}_v \in \mathbb{R}^3$ be the columns
of $\mat{A}_c$ and $\vec{e}_u = (1,0,0)^\top$, $\vec{e}_v = (0,1,0)^\top$.
Differentiating the GS epipolar constraint
with respect to $q_{1,x}$
yields~\cite{Bentolila2014,Barath2018} equation
\begin{equation}\label{eq:ac_gs_u}
  \vec{a}_u^\top \mat{E}\, \vec{q}_1
    + \vec{q}_2^\top \mat{E}\, \vec{e}_u = 0.
\end{equation}
Differentiating with respect to $q_{1,y}$ gives the analogous
\begin{equation}\label{eq:ac_gs_v}
  \vec{a}_v^\top \mat{E}\, \vec{q}_1
    + \vec{q}_2^\top \mat{E}\, \vec{e}_v = 0.
\end{equation}
Together with the epipolar constraint~\eqref{eq:epipolar_gs},
each AC provides \textbf{3 independent equations} on the essential matrix.
Since $\mat{E}$ has 5 degrees of freedom,
just 2 ACs suffice
for essential matrix estimation~\cite{Barath2018}
versus 5 PCs with point-only methods.

\section{RS Relative Pose Estimation}\label{sec:method}

\noindent
Beyond the first-order RS model, we make two explicit
approximations, stated here and justified where used:
\textbf{(i)} constraints are linearized in the motion parameters
$\vec{\theta}$ about $\vec{\theta}=\vec{0}$ (valid as
$\|\vec{\omega}\|<0.15$\,rad, \cref{sec:physical_rs}); and
\textbf{(ii)} the null-space basis eliminating $\vec{\theta}$ is
fixed at the reference rotation $\mat{R}_\mathrm{ref}=\mat{I}$
(\cref{sec:direct_solver}).
The constraints of \cref{sec:ac_rs} are exact; the approx.\
enter only at the solver stage.

\subsection{RS-Corrected Affine Constraints}\label{sec:ac_rs}


The original GS affine constraints~\eqref{eq:ac_gs_u}--\eqref{eq:ac_gs_v}
were derived under the assumption that $\mat{E}$ is constant
across the image.
In the RS case, the essential matrix
$\tilde{\mat{E}}(\tau_1, \tau_2)$ varies with the row coordinates
of both points.
When we perturb $\vec{q}_1$ to derive the affine constraints,
the row coordinates $\tau_1$ and $\tau_2$ change as well,
because a shift in the image position of a point shifts
the row at which it was read out.
This introduces additional terms
$\partial \tilde{\mat{E}} / \partial \tau_k$ that are absent
in the GS formulation.

\vspace{1mm}\noindent\textit{Derivation.}
Consider the RS epipolar constraint
$F(\vec{q}_1, \vec{q}_2) \!=\! \vec{q}_2^\top\, \tilde{\mat{E}}(\tau_1, \tau_2)\, \vec{q}_1 \!=\! 0$.
When we differentiate with respect to $q_{1,x}$ (horizontal shift),
three effects contribute:
(i)~the explicit dependence of $\vec{q}_1$ on its $x$-coordinate
    ($\partial \vec{q}_1/\partial q_{1,x} = \vec{e}_u$);
(ii)~the induced change in $\vec{q}_2$ via the affine map
    ($\partial \vec{q}_2/\partial q_{1,x} = \vec{a}_u$); and
(iii)~the change in the essential matrix caused by
    the vertical shift of $\vec{q}_2$, which alters its readout row $\tau_2$.
Crucially, a horizontal shift of $\vec{q}_1$ does not change
its own readout row $\tau_1$ (rows are horizontal),
but it \emph{may} shift $\vec{q}_2$ vertically through $\vec{a}_u$,
thereby changing $\tau_2$.
Applying the chain rule to all three effects yields
\begin{equation}\label{eq:ac_u}
  \underbrace{\vec{a}_u^\top \tilde{\mat{E}}\, \vec{q}_1
    + \vec{q}_2^\top \tilde{\mat{E}}\, \vec{e}_u}_{\text{same as GS}}
    + \underbrace{\vec{q}_2^\top
      \frac{\partial \tilde{\mat{E}}}{\partial \tau_2}\,
      \vec{q}_1 \cdot \frac{f_y}{h}
      \cdot (\vec{a}_u)_y}_{\text{RS correction}} = 0,
\end{equation}
where $f_y$ is the focal length in pixels, $h$ the image height,
$(\vec{a}_u)_y$ the $y$-component of $\vec{a}_u$
(which converts the image-space vertical shift
to a normalized-row shift via $f_y/h$),
and $\partial\tilde{\mat{E}}/\partial\tau_2$ is given below.

For the vertical derivative ($q_{1,y}$),
a shift of $\vec{q}_1$ now directly changes $\tau_1$,
introducing an additional correction term:
\begin{equation}\label{eq:ac_v}
  \underbrace{\vec{a}_v^\top \tilde{\mat{E}}\, \vec{q}_1
    + \vec{q}_2^\top \tilde{\mat{E}}\, \vec{e}_v}_{\text{same as GS}}
    + \underbrace{\vec{q}_2^\top
      \frac{\partial \tilde{\mat{E}}}{\partial \tau_1}\,
      \vec{q}_1 \cdot \frac{f_y}{h}}_{\substack{\text{RS correction} \\ \text{(from } \tau_1\text{)}}}
    + \underbrace{\vec{q}_2^\top
      \frac{\partial \tilde{\mat{E}}}{\partial \tau_2}\,
      \vec{q}_1 \cdot \frac{f_y}{h}
      \cdot (\vec{a}_v)_y}_{\substack{\text{RS correction} \\ \text{(from } \tau_2\text{)}}} = 0.
\end{equation}
\Cref{eq:ac_v} has \emph{two} correction terms:
a vertical shift of $\vec{q}_1$ changes $\tau_1$ directly
(the $\partial\tilde{\mat{E}}/\partial\tau_1$ term)
and may also shift $\vec{q}_2$ vertically via $(\vec{a}_v)_y$
(the $\partial\tilde{\mat{E}}/\partial\tau_2$ term).

\vspace{1mm}\noindent\textit{Row derivatives of the essential matrix.}
From~\eqref{eq:Rtilde}--\eqref{eq:ttilde},
the derivatives of $\tilde{\mat{R}}$ and $\tilde{\vec{t}}$ with
respect to the row coordinates are
\begin{align}
  \frac{\partial \tilde{\mat{R}}}{\partial \tau_1}
    &= -\bigl(\mat{I} + \tau_2 \skew{\vec{\omega}_2}\bigr)\,
       \mat{R}\,\skew{\vec{\omega}_1}, &
  \frac{\partial \tilde{\vec{t}}}{\partial \tau_1}
    &= -\tilde{\mat{R}}\,\vec{v}_1
       - \tau_1 \frac{\partial \tilde{\mat{R}}}{\partial \tau_1}\,
         \vec{v}_1,
       \label{eq:dRtdtau1} \\
  \frac{\partial \tilde{\mat{R}}}{\partial \tau_2}
    &= \skew{\vec{\omega}_2}\,\mat{R}\,
       \bigl(\mat{I} - \tau_1 \skew{\vec{\omega}_1}\bigr), &
  \frac{\partial \tilde{\vec{t}}}{\partial \tau_2}
    &= \vec{v}_2
       - \tau_1 \frac{\partial \tilde{\mat{R}}}{\partial \tau_2}\,
         \vec{v}_1,
       \label{eq:dRtdtau2}
\end{align}
Combining these with the product rule applied to the
essential matrix~\eqref{eq:Etilde} gives
$\partial\tilde{\mat{E}}/\partial\tau_k
  = \skew{\partial\tilde{\vec{t}}/\partial\tau_k}\;\tilde{\mat{R}}
  + \skew{\tilde{\vec{t}}}\;\partial\tilde{\mat{R}}/\partial\tau_k$.

\vspace{1mm}\noindent\textit{Summary.}
Each AC yields three equations on the RS two-view geometry:
the epipolar constraint~\eqref{eq:epipolar}
plus two new affine constraints~\eqref{eq:ac_u}--\eqref{eq:ac_v}.
The RS correction terms vanish when $\vec{\theta} = \vec{0}$
(since $\partial\tilde{\mat{E}}/\partial\tau_k \to 0$ when there is no RS motion),
recovering the standard GS constraints as a special case.
These constraints are \emph{general}: they hold for any RS two-view
configuration under the first-order RS model
and do not require any linearization.
In the remainder of this section, we exploit them to
build efficient minimal solvers for RANSAC.


\subsection{Minimal Solver from 7 Affine Correspondences}\label{sec:direct_solver}


The RS-corrected AC constraints~\eqref{eq:epipolar},
\eqref{eq:ac_u}--\eqref{eq:ac_v} involve all 17 unknowns
(5 pose + 12 RS motion).
Solving this full nonlinear system directly is algebraically
intractable: even the known-pose subproblem alone (12 RS unknowns)
already has 254 isolated solutions (supplementary material),
and the full 17-DoF system further couples the pose with the
row-dependent $(\tau_1,\tau_2)$ terms.

\vspace{1mm}\noindent\textit{Key observation: the RS parameters enter approximately linearly.}
The epipolar constraint~\eqref{eq:epipolar} is a polynomial
of multi-degree $(2, 1, 2, 1)$ in
$(\vec{\omega}_1, \vec{v}_1, \vec{\omega}_2, \vec{v}_2)$:
it is linear in each translational velocity $\vec{v}_k$
and quadratic in each angular velocity $\vec{\omega}_k$.
However, the RS parameters are physically small
($\|\vec{\omega}_k\| < 0.15$\,rad
for virtually all consumer cameras; see \cref{sec:physical_rs}),
so the quadratic terms
$O(\|\vec{\omega}\|^2) \!<\! 5 \!\times\! 10^{-3}$
are negligible compared to measurement noise.
We therefore \textbf{linearize the RS constraints in $\vec{\theta}$}
around the GS solution $\vec{\theta} = \vec{0}$,
keeping only first-order terms.

For a given pose $(\mat{R}, \vec{t})$,
setting $\vec{\theta} = \vec{0}$ reduces the essential matrix to
$\mat{E}_0 = \skew{\vec{t}}\,\mat{R}$ and the row derivatives
to the compact forms
\begin{equation}\label{eq:dE_domega1}
  \frac{\partial \tilde{\mat{E}}}{\partial \omega_{1,k}}
  \bigg|_{\vec{\theta}=\vec{0}}
  \!=\! -\tau_1 \skew{\vec{t}}\,\mat{R}\,\skew{\vec{e}_k},
  \qquad
  \frac{\partial \tilde{\mat{E}}}{\partial v_{2,k}}
  \bigg|_{\vec{\theta}=\vec{0}}
  \!=\! \tau_2 \skew{\vec{e}_k}\,\mat{R},
\end{equation}
with analogous expressions for $\vec{\omega}_2$ and $\vec{v}_1$.
Each of the three per-AC constraints becomes
a linear function of $\vec{\theta}$ with coefficients that
depend on $(\mat{R}, \vec{t})$ and the data
($\vec{q}_1, \vec{q}_2, \tau_1, \tau_2, \mat{A}_c$).
Stacking all $3N$ linearized constraints for $N$ ACs yields
\begin{equation}\label{eq:linear_system}
  \mat{J}(\mat{R}, \vec{t})\;\vec{\theta}
  = -\vec{r}_0(\mat{R}, \vec{t}),
\end{equation}
where $\vec{r}_0$ collects the GS residuals
(evaluating each constraint at $\vec{\theta}\!=\!0$)
and $\mat{J}$ is the $3N \!\times\! 12$ Jacobian matrix.
The system is {linear in $\vec{\theta}$} but
{nonlinear in $(\mat{R}, \vec{t})$}.

\vspace{1mm}\noindent\textit{Eliminating the RS parameters via null-space projection.}
The key idea of our solver is to \emph{eliminate} the 12
RS unknowns $\vec{\theta}$ from the linear
system~\eqref{eq:linear_system} and obtain conditions
that constrain only the 5 pose unknowns $(\mat{R}, \vec{t})$.

With $N = 7$ ACs, the system~\eqref{eq:linear_system}
has $3 \times 7 = 21$ rows and 12 columns.
It is overdetermined in $\vec{\theta}$:
there are more equations than unknowns.
Such a system has a solution only if the right-hand side
$-\vec{r}_0$ lies in the column space of $\mat{J}$.
Equivalently, $\vec{r}_0$ must be orthogonal to every vector
in the left null space of $\mat{J}$.
Let $\mat{U}_\perp \in \mathbb{R}^{21 \times 9}$ be an
orthonormal basis for this $21 - 12 = 9$-dimensional subspace.
Then~\eqref{eq:linear_system} has a solution if and only if
\begin{equation}\label{eq:consistency}
  \mat{U}_\perp^\top \vec{r}_0(\mat{R}, \vec{t}) = \vec{0}.
\end{equation}
This projects out the 12 RS unknowns entirely,
yielding \textbf{9 polynomial equations in the 5 pose unknowns}
$(\mat{R}, \vec{t})$ alone.
Once a pose satisfying~\eqref{eq:consistency} is found,
the RS parameters are recovered from~\eqref{eq:linear_system}
by least squares fitting.

\vspace{1mm}\noindent\textit{Fixed null-space basis.}
Strictly, $\mat{U}_\perp$ depends on $(\mat{R}, \vec{t})$
through $\mat{J}$, which would make~\eqref{eq:consistency}
implicitly nonlinear in the pose.
We avoid this complication by computing $\mat{U}_\perp$ once
at a fixed reference rotation
$\mat{R}_\mathrm{ref} = \mat{I}$ (identity)
and a generic translation.
This is an approximation, justified by two observations.
First, $\mat{J}$ depends on the pose only through
the linearization point of the RS model,
and its left null space varies smoothly -- so
moderate deviations from $\mat{I}$ preserve
the rank structure.
Second, we verify empirically that
$\mat{R}_\mathrm{ref} = \mat{I}$
yields equal or better numerical conditioning than
recomputing $\mat{U}_\perp$ at the initial 5-point estimate,
producing a success rate above 85\%
for rotation errors below $1^\circ$ on synthetic data.
A pose-dependent $\mat{U}_\perp$ would moreover break the
linear-in-$\vec{t}$ factorization that yields the clean
$\mat{G}_\mathrm{poly}$ structure of~\eqref{eq:G_system}.

\vspace{1mm}\noindent\textit{Polynomial formulation.}
The 9 consistency conditions~\eqref{eq:consistency} involve
the rotation $\mat{R}$ and translation $\vec{t}$,
which we now parameterize to obtain a polynomial system.
We use the Cayley parameterization
$\mat{R}(\vec{s}) = (\mat{I} - \skew{\vec{s}})(\mat{I} + \skew{\vec{s}})^{-1}$,
where $\vec{s} = (s_1, s_2, s_3) \in \mathbb{R}^3$,
and write the translation as $\vec{t} = (1, u, v)^\top$ (up to scale;
this assumes $t_x \neq 0$, which holds generically
and can be ensured by a coordinate permutation
when $|t_x|$ is small).

A crucial simplification comes from the fact that
$\vec{r}_0$ depends \emph{linearly} on $\vec{t}$.
This means~\eqref{eq:consistency} can be factored as
\begin{equation}\label{eq:G_system}
  \mat{G}_\mathrm{poly}(\vec{s})
  \begin{pmatrix}1 \\ u \\ v\end{pmatrix} = \vec{0},
\end{equation}
where $\mat{G}_\mathrm{poly}(\vec{s})$ is a $9 \times 3$ matrix
whose entries are degree-2 polynomials in $\vec{s}$.
(The Cayley parameterization introduces a common denominator
$d(\vec{s}) = 1 + \|\vec{s}\|^2$,
which we absorb into $\mat{G}_\mathrm{poly}$
to clear fractions.)
We determine $\mat{G}_\mathrm{poly}$ numerically:
each entry has 10 monomials
$\{1, s_1, s_2, s_3, s_1^2, s_1 s_2, \ldots, s_3^2\}$,
so evaluating~\eqref{eq:consistency} at 20 random
$\vec{s}$-samples and solving a $20 \times 10$
least-squares system per entry recovers the polynomial coefficients
(to machine precision).

\vspace{1mm}\noindent\textit{Eliminating the translation via minors.}
The system~\eqref{eq:G_system} states that
$(1, u, v)^\top$ lies in the null space of the $9 \times 3$ matrix
$\mat{G}_\mathrm{poly}(\vec{s})$.
A nontrivial null vector exists if and only if
$\mat{G}_\mathrm{poly}$ has rank at most 2, \ie
all its $3 \times 3$ subdeterminants (minors) vanish.
To work with a smaller set of minors,
we first reduce from 9 rows to 5 by forming
5 random linear combinations
(with i.i.d\onedot\ Gaussian coefficients),
yielding a $5 \times 3$ matrix
$\hat{\mat{G}}(\vec{s})$ with degree-2 entries.
By B\'ezout's theorem~\cite{CoxLittleOShea2015},
the solution set is independent of the (generic) choice
of combination coefficients.
We reduce to $5$ rows rather than fewer because $5\times3$ is the
smallest reduction whose minor ideal is zero-dimensional:
a $4\times3$ reduction leaves a positive-dimensional solution set
in $\vec{s}$.
We then require all
$\binom{5}{3} = 10$ maximal minors of $\hat{\mat{G}}$ to vanish.
Each $3 \times 3$ determinant has degree $3 \times 2 = 6$ in $\vec{s}$.
A key algebraic observation simplifies this system:
every minor is divisible by the Cayley denominator
$d(\vec{s}) = 1 + \|\vec{s}\|^2$.
This occurs because $\mat{G}_\mathrm{poly}$ was constructed
by multiplying the rational matrix $\mat{G}(\vec{s})$ by $d(\vec{s})$;
at the complex roots of $d(\vec{s}) = 0$,
the Cayley rotation is undefined (it corresponds to a $180^\circ$ rotation),
and the original constraints impose no condition on $(u,v)$,
so $\hat{\mat{G}}$ must drop rank there,
forcing $d(\vec{s})$ to divide every maximal minor.
We verify this computationally by evaluating
each minor at random complex points on the variety
$\{d(\vec{s}) = 0\}$, confirming vanishing
to machine precision ($\sim\! 10^{-14}$),
while $d(\vec{s})^2$ does \emph{not} divide the minors.

After dividing out $d(\vec{s})$, we obtain
{10 polynomial equations of degree 4}
in $\vec{s} = (s_1, s_2, s_3)$.
The number of isolated solutions is determined by the
Hilbert function~\cite{CoxLittleOShea2015} of the ideal,
which stabilizes at $H = 20$
(verified by computing the rank of Macaulay matrices
at successive degrees).
This establishes the {algebraic degree of the system as 20}.

\vspace{1mm}\noindent\textit{Action matrix solver.}
We extract all 20 solutions using action
matrices~\cite{CoxLittleOShea2015}, a standard technique
in computational algebraic geometry for solving
zero-dimensional polynomial systems.
The idea is to construct a matrix whose eigenvalues
are the coordinates of the solutions.

We build the Macaulay matrix at extension degree
$d_\mathrm{ext} = 5$, which has dimensions $40 \times 56$.
This matrix is formed by multiplying each of the 10 degree-4
polynomials by the 4 monomials $\{1, s_1, s_2, s_3\}$
(degree $\leq 1$), producing 40 rows;
columns index all $\binom{5+3}{3} = 56$ monomials of
degree $\leq 5$ in $(s_1, s_2, s_3)$.
The SVD of this matrix reveals a clear gap at the 36th singular value
(ratio $\sigma_{36}/\sigma_{37} > 10^3$),
confirming a 20-dimensional null space
$\mat{V}_\mathrm{null} \in \mathbb{R}^{56 \times 20}$
that represents the quotient algebra of the ideal.
For each variable $s_k$, the action matrix is
\begin{equation}\label{eq:action}
  \mat{M}_k = \mat{V}_\mathrm{null}^\top\, \mat{S}_k\, \mat{V}_\mathrm{null}
  \in \mathbb{R}^{20 \times 20},
\end{equation}
where $\mat{S}_k$ is the $56 \times 56$ monomial shift matrix
that maps each basis monomial $m$ to $s_k \cdot m$
(or to zero if $\deg(s_k \cdot m) > 5$).
The eigenvalues of $\mat{M}_k$ are precisely
the $s_k$-coordinates of the 20 solutions.
Since the three action matrices commute,
they can be simultaneously diagonalized:
eigendecomposing $\mat{M}_1$ and applying the same
eigenvector transform to $\mat{M}_2$ and $\mat{M}_3$
yields the complete triples
$(s_1^{(i)}, s_2^{(i)}, s_3^{(i)})$ for $i = 1, \ldots, 20$.

\vspace{1mm}\noindent\textit{Solution recovery.}
For each real Cayley vector $\vec{s}^{(i)}$,
the translation parameters $(u, v)$ are recovered from
the overdetermined system~\eqref{eq:G_system} via least squares,
and the pose is computed as
$\mat{R} = (\mat{I} - \skew{\vec{s}})(\mat{I} + \skew{\vec{s}})^{-1}$,
$\vec{t} = (1, u, v)^\top / \|(1, u, v)\|$.
Solutions are ranked by the polynomial residual
$\|\mat{G}_\mathrm{poly}(\vec{s})(1, u, v)^\top\|$,
which consistently identifies the correct solution.
The RS parameters $\vec{\theta}$ are then recovered
from~\eqref{eq:linear_system} by least squares.

\vspace{1mm}\noindent\textit{Computational cost.}
The algebraic solver runs in ${\sim}1.2$\,ms in C++ (null space
$0.04$, $\mat{G}_\mathrm{poly}$ fitting $0.2$, minors and $d(\vec{s})$
division $0.02$, $40\times56$ SVD $0.3$, action matrices and
eigendecomposition $0.6$\,ms).
Though slower than the microsecond solvers ideal for RANSAC inner
loops, the reduced sample size ($k{=}7$ vs.\ $20$-$44$) more than
compensates via far fewer iterations.

\vspace{1mm}\noindent\textit{Degeneracies.}
Null-space elimination requires $\mat{J}$ to have full column rank
(12); this fails for degenerate samples (\eg all 7 ACs on one row,
where the $\tau$-dependent terms vanish).
Such cases are rare in RANSAC and are detected via the SVD gap of the
$40\times56$ Macaulay matrix -- if absent, the solver returns nothing
and RANSAC proceeds.

\vspace{1mm}\noindent\textit{Remark: 6-AC solver.}
Each AC gives 3 constraints, so $N=5$ ACs underdetermine the 17-DoF
system while $N=6$ overdetermine it by one, making 6 ACs
theoretically sufficient.
However, the resulting $6\times3$ matrix is too small for the
minor-based elimination above; a hidden-variable
alternative~\cite{Kukelova2008} instead suffers from a
positive-dimensional spurious component (46-dimensional null space)
that destabilizes extraction and costs ${\sim}55$\,ms per call --
unsuitable for RANSAC (full derivation in the supplementary material).


\subsection{Integration into RANSAC}\label{sec:ransac}


We integrate the 7-AC solver into
SupeRANSAC~\cite{barath2025superansac}
with local optimization.

\vspace{1mm}\noindent\textit{Hypothesis generation.}
For each RANSAC iteration, we draw 7 ACs and run:
(1)~Nist\'er's 5-point algorithm~\cite{Nister2004} on the
first 5 point coordinates, selecting the best $\mat{E}$ via
the 6th and 7th ACs and decomposing into $(\mat{R}, \vec{t})$;
then (2)~the action-matrix solver (\cref{sec:direct_solver}),
producing up to 20 candidates filtered by cheirality
and physical RS parameter bounds.
A GS fallback ($\vec{\theta} = \vec{0}$) is always included.
Models are scored via the RS Sampson distance~\cite{Dai2016}
  $d_S^2 = ((\vec{q}_2^\top \tilde{\mat{E}}_i\, \vec{q}_1)^2) /
    (\|\tilde{\mat{E}}_i\, \vec{q}_1\|_{1:2}^2
    + \|\tilde{\mat{E}}_i^\top \vec{q}_2\|_{1:2}^2)$,
where $\tilde{\mat{E}}_i = \tilde{\mat{E}}(\tau_{1,i}, \tau_{2,i})$
is the essential matrix of pair $i$.

\vspace{1mm}\noindent\textit{Non-minimal refinement.}
The best hypotheses are refined via joint
Levenberg--Marquardt (LM) over all 17 parameters on the inlier set,
minimizing
\begin{equation}\label{eq:lm_cost}
  C = \sum_{i=1}^{N_\mathrm{in}}
    \bigl(r_{0,i}^2 + w_\mathrm{ac}^2\, r_{1,i}^2 + w_\mathrm{ac}^2\, r_{2,i}^2\bigr),
\end{equation}
where $r_{0,i}$ is the epipolar residual~\eqref{eq:epipolar}
and $r_{1,i}, r_{2,i}$ the affine
residuals~\eqref{eq:ac_u}--\eqref{eq:ac_v} of pair $i$, each
evaluated with its own $\tilde{\mat{E}}_i$ under the full nonlinear
RS model.
All three residuals are computed on normalized image coordinates
($\mat{K}^{-1}\vec{q}$) and thus share a common scale;
we keep the affine terms in algebraic form since, unlike the
epipolar residual, they have no direct point-to-line distance analogue.
The weight $w_\mathrm{ac}$ balances the affine against the epipolar
terms: all reported results use $w_\mathrm{ac}=1$, and the pose AUC is
insensitive to this choice for $w_\mathrm{ac}\in[0.5,2]$;
a principled, data-driven alternative sets $w_\mathrm{ac}$ to the
ratio of the held-out standard deviations of the epipolar and
affine residuals.

\vspace{1mm}\noindent\textit{Handling the $\vec{v}$-$\vec{t}$ coupling.}
The translational velocity $\vec{v}_k$ is poorly separated from
$\vec{t}$ through epipolar constraints
alone~\cite{AitAider2009,Dai2016}, making the LM Jacobian ill-conditioned
along $\vec{t}$.
We apply mild damping $\lambda_v \mat{I}$ on the $\vec{v}$-blocks,
biasing toward small velocities.
The affine residuals additionally help alleviate this coupling,
as they depend on $\partial\tilde{\mat{E}}/\partial\tau_k$
and thus provide information to distinguish $\vec{v}_k$ from $\vec{t}$:
while epipolar constraints measure only the projection
of motion along epipolar lines -- where $\vec{v}$ and $\vec{t}$
produce nearly identical effects --
the AC constraints capture the \emph{spatial gradient}
of this projection across rows,
which varies differently with $\vec{v}$
(row-dependent) than with $\vec{t}$ (constant).


\subsection{Physically Feasible RS Parameters}\label{sec:physical_rs}


The RS motion parameters $\vec{\omega}_k$ and $\vec{v}_k$
represent the \emph{total} rotation and translation accumulated
during the sensor readout,
so their physical magnitudes are tightly constrained.
Since the model parameters relate to their physical counterparts as
$\|\vec{\omega}\|_\mathrm{model} = \|\vec{\omega}\|_\mathrm{phys} \cdot t_\mathrm{ro}$
and
$\|\vec{v}\|_\mathrm{model} = \|\vec{v}\|_\mathrm{phys} \cdot t_\mathrm{ro}$,
even an aggressively maneuvering drone
(angular velocity $\sim$10\,rad/s, translational velocity $\sim$30\,m/s)
with a typical readout time $t_\mathrm{ro} \approx 30$\,ms
yields $\|\vec{\omega}\| \leq 0.3$\,rad ($\sim$17$^\circ$) and
$\|\vec{v}\| \leq 0.9$.
For handheld cameras, the values are 1--2 orders of magnitude
smaller.
Notably, while $\|\vec{\omega}\|$ remains small across all scenarios,
$\|\vec{v}\|$ can reach ${\sim}0.9$\,m for fast-moving platforms,
underscoring the importance of accurately estimating $\vec{v}$ --
which, as we show in \cref{sec:exp_synthetic},
is poorly conditioned from point correspondences alone.
This has two important consequences for our solver.

\vspace{1mm}\noindent\textit{Justification of linearization.}
Since the RS parameters are small, the nonlinear terms
in the epipolar constraint -- which scale as
$\tau^2 \|\vec{\omega}\|^2 \leq 0.25 \cdot 0.02 = 5 \times 10^{-3}$ -- are
2--3 orders of magnitude below typical measurement noise.
The linearization in $\vec{\theta}$ used
by our solver (\cref{sec:direct_solver}) is therefore
an excellent approximation,
not merely a practical heuristic.

\vspace{1mm}\noindent\textit{Early rejection.}
The physical constraints on $\vec{\theta}$ provide a useful
sanity check during RANSAC: candidate models with
$\|\vec{\omega}\| > 0.5$\,rad or unreasonably large $\|\vec{v}\|$
can be discarded, avoiding unnecessary
scoring over the full correspondence set.

\section{Experiments}\label{sec:experiments}

We evaluate the proposed RS-7AC solver in three settings:
(i)~controlled synthetic experiments that isolate the effect
of individual noise sources on solver accuracy,
(ii)~real-world experiments on the TUM RS
dataset~\cite{Schubert2018} within a full RANSAC pipeline,
and (iii)~a test on the global-shutter
EuRoC MAV dataset~\cite{Burri2016} -- the special case of zero
readout time -- to verify the method does
not degrade in the absence of RS distortion.
We compare against
GS-5PC (Nist\'er's 5-point algorithm~\cite{Nister2004},
which ignores RS),
GS-2AC (the 2-AC essential matrix solver of
Barath and Hajder~\cite{Barath2018},
which also ignores RS but uses affine correspondences),
and the RS-20PC and RS-44PC solvers of
Dai et al.~\cite{Dai2016} (linearized and GN-refined).


\subsection{Synthetic Experiments}\label{sec:exp_synthetic}


We generate synthetic RS two-view scenes (50 random 3D points;
two RS cameras with $f_x\!=\!f_y\!=\!500$\,px, $640\times480$;
random relative pose with rotation $\leq10^\circ$ and forward-dominant
translation; RS motion scaled by a factor $\alpha$ applied to both
$\vec{\omega}_k$ and $\vec{v}_k$).
Ground-truth affine frames follow from central-finite-difference
differentiation of the full RS projection (including the iterative
row solve): $\mat{A}_c = (\partial\vec{q}_2/\partial\vec{X})
(\partial\vec{q}_1/\partial\vec{X})^{\dagger}$.

Each solver receives its minimal sample
(5/20/44 PCs for GS-5PC/RS-20PC/RS-44PC, 7 ACs for RS-7AC)
with oracle model selection; since minimal models are too noisy to
compare directly, the best is refined on all 50 correspondences by
joint LM, measuring each solver's quality as a RANSAC initialization.
RS-7AC refines with the combined cost~\eqref{eq:lm_cost}
($w_\mathrm{ac}=1$), the others with epipolar residuals only.
We report median errors over 500 trials.

We sweep
(1)~point noise $\sigma_p \in [0, 3]$\,px (fixed $\alpha = 0.5$),
(2)~affine noise $\sigma_a \in [0, 0.1]$ (fixed $\sigma_p = 0.5$\,px, $\alpha = 0.5$), and
(3)~RS magnitude $\alpha \in [0, 3]$ (fixed $\sigma_p = 0.5$\,px, $\sigma_a = 0.01$),
where $\alpha$ scales both $\vec{\omega}_k$ and $\vec{v}_k$ simultaneously.
Affine noise is additive Gaussian on each entry
of the $3 \times 2$ affine matrix.
The swept range is calibrated to real data:
$\sigma_a \approx 0.05$ matches the median AC residual of AffNet and
$\sigma_a \approx 0.1$ that of RoMa on TUM-RS (supplementary material),
so the tested noise levels reflect those of practical matchers.

\begin{figure}[t]
\centering
\includegraphics[width=0.99\linewidth]{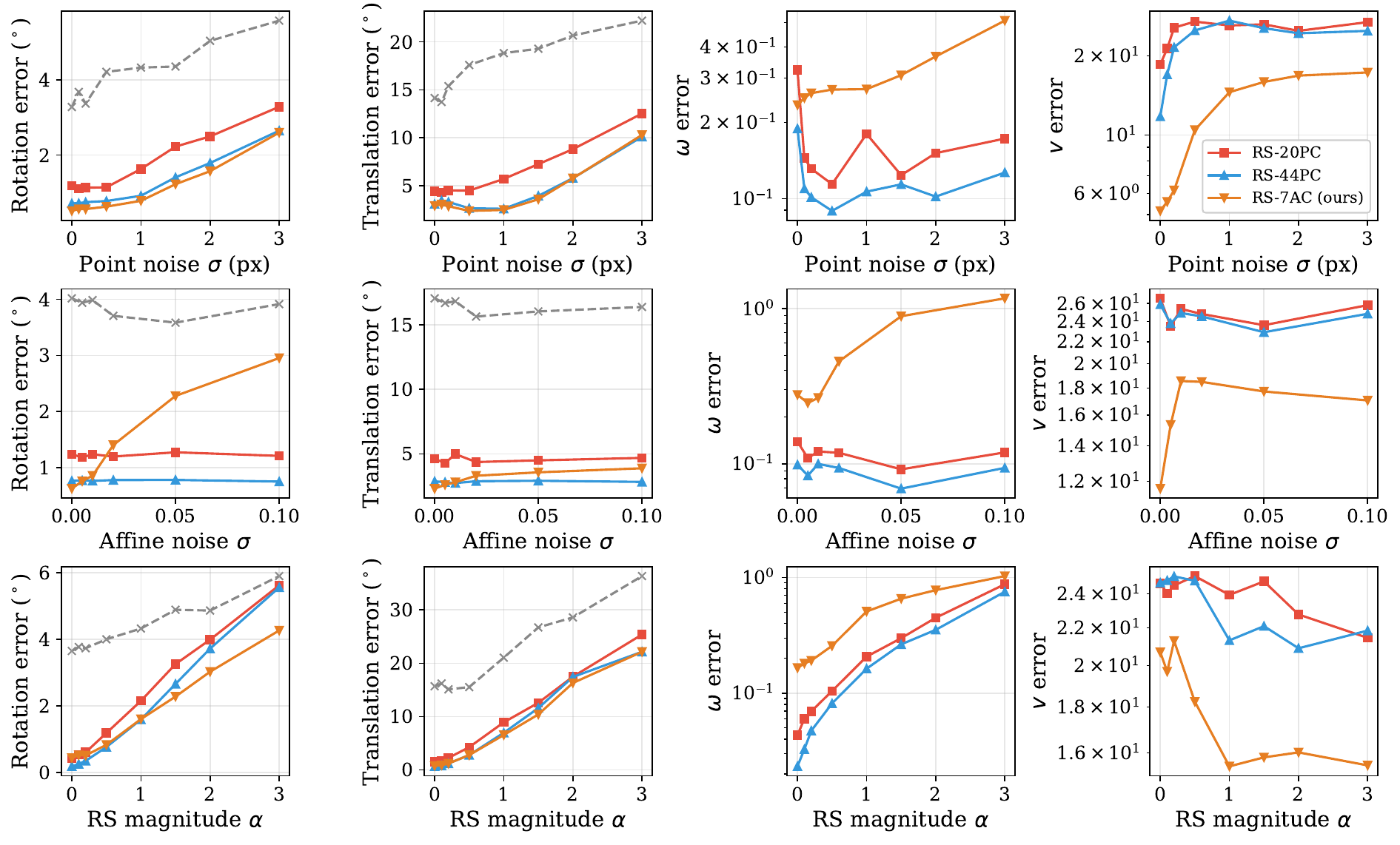}
\caption{Synthetic solver evaluation.
Median errors over 500 trials for four metrics (rows:
rotation, translation, $\vec{\omega}$ error, $\vec{v}$ error)
as a function of three noise parameters
(columns: point noise, affine noise, RS magnitude~$\alpha$).
GS-5PC is excluded from the RS parameter rows as it does not estimate
RS parameters.
RS-7AC achieves pose accuracy competitive with RS-44PC
while using $6\times$ fewer correspondences,
and is the only method with meaningful translational
velocity estimates.}\label{fig:synthetic}

\end{figure}

\Cref{fig:synthetic} shows the results.
For \emph{pose estimation} (top two rows),
RS-7AC achieves competitive pose accuracy
despite using only 7 correspondences versus 20-44.
At moderate noise ($\sigma_p = 0.5$\,px, $\alpha = 0.5$),
the median rotation error is $0.65^\circ$ for RS-7AC
versus $0.46^\circ$ for RS-44PC and $4.21^\circ$ for GS-5PC.
RS-44PC benefits from its larger sample in this oracle setting,
but the gap is small and, as we show in \cref{sec:exp_real},
is more than offset by the RANSAC advantage
of the smaller sample size in real-world conditions.
The RS-aware methods degrade gracefully with increasing RS magnitude,
whereas GS-5PC, which ignores RS, accumulates systematic bias.
For \emph{RS parameter estimation} (bottom two rows),
RS-7AC exhibits ${\sim}2\times$ higher angular velocity error
$\vec{\omega}$ than the point-based baselines,
as the 7-AC algebraic solver provides a less accurate $\vec{\omega}$ initialization
than the overdetermined linear systems of RS-20PC and RS-44PC.
However, the translational velocity $\vec{v}$ reveals the
opposite trend:
RS-20PC and RS-44PC, which rely solely on point correspondences,
suffer from the $\vec{v}$-$\vec{t}$ coupling
discussed in \cref{sec:ransac},
yielding median velocity errors of $\sim$25 --
far worse than predicting zero.
RS-7AC, leveraging affine constraints in both estimation
and refinement, achieves ${\sim}3\times$ lower velocity error ($\sim$10),
confirming that ACs provide information
about $\vec{v}$ that PCs alone cannot resolve.

A direct comparison of raw solver outputs is not meaningful here,
as the methods consume different inputs (7 ACs vs.\ 20 or 44 PCs);
the oracle setup instead measures each solver's quality as an
initialization for refinement.
Under real RANSAC with outliers (\cref{tab:pose_auc}),
RS-7AC outperforms the baselines on \emph{both} $\vec{\omega}$ and
$\vec{v}$, as the large samples of RS-20PC and RS-44PC cripple
clean-sample selection.


\subsection{Real-World Experiments}\label{sec:exp_real}


We evaluate on the TUM-RS dataset~\cite{Schubert2018}
(10 RS fisheye sequences with motion-capture ground truth),
testing all 10 sequences at strides of 10 and 20 frames
(increasing baseline and RS effect).
We extract ACs with two feature types:
(i)~\textbf{Key.Net~\cite{Laguna2019} + AffNet~\cite{Mishkin2018} + HardNet~\cite{Mishchuk2017}},
a classical pipeline (Key.Net keypoints, AffNet local affine shape,
HardNet descriptor, mutual-NN ratio matching), with $\mat{A}_c$ from
the matched affine shapes; and
(ii)~\textbf{RoMa~\cite{Edstedt2024}}, a dense matcher whose
pixel-level warp field we differentiate (central finite differences)
to get a per-point $2\times2$ Jacobian.
Since every warp-field sample yields one AC, $\#$ACs $=$ $\#$PCs at no
extra cost -- ACs are thus not fewer than PCs for dense matchers;
their per-point accuracy is quantified in the supplementary material.
All methods run in the same
SupeRANSAC~\cite{barath2025superansac} pipeline
(PROSAC sampling~\cite{chum2005matching}, MAGSAC scoring~\cite{Barath2020MAGSACpp},
nested local optimization).
We report pose AUC at $5^\circ$, $10^\circ$, and $20^\circ$,
the per-pair error being $\max(\epsilon_R,\epsilon_t)$.

\begin{table}[t]
\centering
\caption{\textbf{Pose and RS parameter estimation on the TUM-RS dataset~\cite{Schubert2018}.}
All methods use SupeRANSAC~\cite{barath2025superansac} with
MSAC scoring and 5\,000 iterations.
Pose accuracy is measured by AUC of the cumulative error curve
at $5^\circ$, $10^\circ$, and $20^\circ$ thresholds,
where the error of each pair is $\max(\epsilon_R, \epsilon_t)$.
RS parameter accuracy is the median $\ell_2$ error of the
angular ($\epsilon_\omega$, rad) and translational
($\epsilon_v$, m) velocities summed over both cameras.
GS-5PC and GS-2AC do not estimate RS parameters (--).
Two feature extractors are compared:
AffNet~\cite{Mishkin2018} (sparse, ${\sim}$500 ACs)
and RoMa~\cite{Edstedt2024} (dense, ${\sim}$10\,000 ACs).
\textbf{Bold} marks the best result per column and extractor.}\label{tab:pose_auc}

\resizebox{\linewidth}{!}{%
\begin{tabular}{cl|ccccc|ccccc|c}
\toprule
 & & \multicolumn{5}{c|}{stride = 10} & \multicolumn{5}{c|}{stride = 20} & \multirow{2}{*}{$t$ (secs) $\downarrow$} \\
& Method & AUC@5$^\circ$ $\uparrow$ & @10$^\circ$ $\uparrow$ & @20$^\circ$ $\uparrow$ & $\epsilon_\omega$ $\downarrow$ & $\epsilon_v$ $\downarrow$ & AUC@5$^\circ$ $\uparrow$ & @10$^\circ$ $\uparrow$ & @20$^\circ$ $\uparrow$ & $\epsilon_\omega$ $\downarrow$ & $\epsilon_v$ $\downarrow$ &  \\
\midrule
\multirow{5}{*}{\rotatebox{90}{AffNet}} & RS-20PC~\cite{Dai2016} & 0.101 & 0.181 & 0.258 & 0.177 & 11.260 & 0.097 & 0.161 & 0.228 & 0.183 & 8.861 & 12.7  \\
 & RS-44PC~\cite{Dai2016}  & 0.267 & 0.366 & 0.451 & 0.097 & 13.110 & 0.240 & 0.319 & 0.389 & 0.102 & 10.752 & 25.2 \\
 & GS-5PC~\cite{Nister2004} & 0.428 & 0.535 & 0.626 & -- &  -- & 0.380 & 0.472 & 0.544 & -- & -- & \phantom{1}1.5 \\
 & GS-2AC~\cite{Barath2018} & 0.309 & 0.490 & 0.660 & -- & -- & 0.288 & 0.447 & 0.600 & -- & -- & \phantom{1}\textbf{0.3} \\
 & Proposed & \textbf{0.502} & \textbf{0.609} & \textbf{0.687} & \textbf{0.047} & \textbf{0.056}  & \textbf{0.476} & \textbf{0.573} & \textbf{0.645} & \textbf{0.045} & \textbf{0.054} & \phantom{1}3.7 \\
\midrule
 \multirow{5}{*}{\rotatebox{90}{RoMA}} & RS-20PC~\cite{Dai2016} & 0.769 & 0.869 & 0.924 & 0.063 & 8.011 & 0.687 & 0.773 & 0.826 & 0.071 & 5.595 & 20.9 \\
 & RS-44PC~\cite{Dai2016} & 0.870 & 0.929 & 0.960 & 0.050 & 9.537 & 0.772 & 0.827 & 0.857 & 0.050 & 7.721 & 46.3  \\
 & GS-5PC~\cite{Nister2004} & 0.861 & 0.908 & 0.934 & -- & -- & 0.757 & 0.803 & 0.830 & -- & -- & \phantom{1}5.3 \\
 & GS-2AC~\cite{Barath2018} & 0.859 & 0.926 & 0.962 & -- & -- & 0.772 & 0.838 & 0.876 & -- & -- & \phantom{1}\textbf{1.3 }\\
 & Proposed & \textbf{0.897} & \textbf{0.948} & \textbf{0.973} & \textbf{0.043} & \textbf{0.051} & \textbf{0.810} & \textbf{0.860} & \textbf{0.886} & \textbf{0.044} & \textbf{0.053}  & \phantom{1}7.5  \\
\bottomrule
\end{tabular}%
}

\end{table}

\Cref{tab:pose_auc} shows the results.
The proposed RS-7AC solver consistently outperforms all baselines
by a significant margin.
At stride~10, RS-7AC achieves an AUC@$5^\circ$ of 0.502,
improving over GS-5PC (0.428) by 17\% relative
and over RS-44PC (0.267) by 88\%.
The advantage persists at stride~20,
where the larger baseline amplifies the RS effect.

The strong performance of the GS methods over the Dai et~al.\ solvers
stems from sample size: drawing an all-inlier sample of size $k$ at
outlier ratio $\epsilon$ has probability $(1-\epsilon)^k$, which at
$\epsilon=30\%$ drops from $0.08$ for $k=7$ to $\sim\!10^{-7}$ for
$k=44$, crippling RANSAC for the 20- and 44-PC solvers.
GS-2AC~\cite{Barath2018} benefits from an even smaller sample
($k{=}2$) but lacks RS modeling.
Our 7-AC sample thus combines small-sample RANSAC efficiency with
RS-aware modeling.

\vspace{1mm}\noindent\textit{Runtime.}
RS-20PC and RS-44PC are by far the slowest
(13--46\,s) due to their large samples;
RS-7AC runs in 3.7\,s (AffNet) and 7.5\,s (RoMa) -- only
${\sim}2\times$ slower than GS-5PC despite a 17-DoF vs.\ 5-DoF problem,
thanks to its small sample.
GS-2AC is fastest (${\sim}0.3$-$1.3$\,s) but lacks RS modeling.

\vspace{1mm}\noindent
\textbf{Global-Shutter Special Case.}
%
To verify that the RS solver does not degrade without RS distortion,
we evaluate on the EuRoC MAV dataset~\cite{Burri2016}
(six indoor global-shutter sequences, 752$\times$480, 20\,fps, three
from Vicon Room~1 and three from Machine Hall, with Vicon ground truth),
using RoMa features at strides 10 and 20.
As the camera is global-shutter, the RS parameters should be
near-zero, and pose accuracy measures how gracefully the method
handles their absence.

\begin{table}[t]
\centering
\caption{\textbf{Pose estimation on EuRoC MAV~\cite{Burri2016} (global-shutter cameras).}
Same setup as \cref{tab:pose_auc} but on GS images.
RS parameters should be near-zero; $\epsilon_\omega$ and $\epsilon_v$
measure how well each RS solver handles the GS case.
\textbf{Bold} marks the best result per column.
The proposed solver achieves comparable performance to GS solvers.}\label{tab:euroc_auc}

\resizebox{0.9\linewidth}{!}{%
\begin{tabular}{l|ccccc|ccccc}
\toprule
 & \multicolumn{5}{c|}{stride = 10} & \multicolumn{5}{c}{stride = 20} \\
Method & AUC@5$^\circ$ $\uparrow$ & @10$^\circ$ $\uparrow$ & @20$^\circ$ $\uparrow$ & $\epsilon_\omega$ $\downarrow$ & $\epsilon_v$ $\downarrow$ & AUC@5$^\circ$ $\uparrow$ & @10$^\circ$ $\uparrow$ & @20$^\circ$ $\uparrow$ & $\epsilon_\omega$ $\downarrow$ & $\epsilon_v$ $\downarrow$ \\
\midrule
RS-20PC~\cite{Dai2016} & 0.426 & 0.631 & 0.765 & 0.075 & 11.762 & 0.421 & 0.580 & 0.692 & 0.072 & 11.785 \\
RS-44PC~\cite{Dai2016}  & 0.499 & 0.687 & 0.800 & 0.041 & 19.296 & 0.467 & 0.622 & 0.726 & 0.035 & 14.763 \\
GS-5PC~\cite{Nister2004} & \textbf{0.521} & \textbf{0.704} & 0.813 & -- &  -- & \textbf{0.492} & 0.644 & 0.741 & -- & -- \\
GS-2AC~\cite{Barath2018} & 0.517 & 0.702 & 0.814 & -- & -- & 0.491 & \textbf{0.647} & \textbf{0.749} & -- & -- \\
Proposed & 0.512 & 0.701 & \textbf{0.816} & \textbf{0.016} & \textbf{0.035}  & 0.484 & 0.641 & 0.744 & \textbf{0.012} & \textbf{0.028} \\
\bottomrule
\end{tabular}%
}

\end{table}

\Cref{tab:euroc_auc} shows the results.
The proposed solver achieves similar pose accuracy to the GS baselines,
demonstrating that it gracefully handles the GS special case.
Notably, it correctly recovers near-zero RS parameters
($\epsilon_\omega \approx 0.01$\,rad, $\epsilon_v \approx 0.03$),
confirming that the solver does not introduce spurious distortions
when RS effects are absent.
In contrast, the Dai et~al.\ solvers yield much higher RS parameter errors
($\epsilon_\omega \approx 0.04$-$0.08$\,rad, $\epsilon_v \approx 10$-$20$),
as their large minimal samples make RANSAC less likely
to find clean inlier sets even in the GS regime.

\section{Conclusion}\label{sec:conclusion}

We introduced affine correspondences into rolling shutter
two-view geometry and developed a linearized 7-AC solver
(degree~20, 1.8\,ms) that reduces the RANSAC sample
from 20 to~7, making RS-aware robust estimation practical.
On TUM-RS, the method achieves the best pose accuracy
among all baselines and substantially improves
translational velocity estimation, which is poorly conditioned
from point correspondences alone.
On the global-shutter EuRoC MAV dataset, the solver
handles the GS special case gracefully, being comparable to the GS 5-point algorithm
while correctly recovering near-zero RS parameters.

\bibliographystyle{splncs04}
\bibliography{main}

\end{document}